\documentclass[conference]{IEEEtran}
\IEEEoverridecommandlockouts
\usepackage{cite}
\usepackage{amsmath,amssymb,amsfonts}
\usepackage{algorithmic}
\usepackage{graphicx}
\usepackage{textcomp}
\usepackage{xcolor}
\usepackage{array}
\usepackage{makecell}
\def\BibTeX{{\rm B\kern-.05em{\sc i\kern-.025em b}\kern-.08em
    T\kern-.1667em\lower.7ex\hbox{E}\kern-.125emX}}
\begin{document}

\title{A Hybrid Transformer Model for Fake News Detection: Leveraging Bayesian Optimization and Bidirectional Recurrent Unit
	\thanks{$^\ast$ Corresponding author: tianyihuang@berkeley.edu}
%	\thanks{$^\dagger$ These authors contributed equally to this work.}
}

\author{
\IEEEauthorblockN{1\textsuperscript{st} Tianyi Huang\textsuperscript{$\ast$}}
\IEEEauthorblockA{\textit{Department of Electrical Engineering and Computer Sciences} \\
	\textit{University of California}\\
	Berkeley, CA 94720 \\
	tianyihuang@berkeley.edu}
\and 
\IEEEauthorblockN{1\textsuperscript{st} Zeqiu Xu}
\IEEEauthorblockA{\textit{Information Networking Institute} \\
	\textit{Carnegie Mellon University}\\
	Pittsburgh, PA 15213 \\
	zeqiux@alumni.cmu.edu }
\and 
\IEEEauthorblockN{2\textsuperscript{nd} Peiyang Yu}
\IEEEauthorblockA{\textit{Information Networking Institute} \\
	\textit{Carnegie Mellon University}\\
	Pittsburgh, PA 15213 \\
	peiyangy@alumni.cmu.edu}
\and 
\IEEEauthorblockN{2\textsuperscript{nd} Jingyuan Yi}
\IEEEauthorblockA{\textit{Information Networking Institute} \\
	\textit{Carnegie Mellon University}\\
	Pittsburgh, PA 15213 \\
	jingyuay@alumni.cmu.edu}
\and 
\IEEEauthorblockN{3\textsuperscript{th} Xiaochuan Xu}
\IEEEauthorblockA{\textit{Information Networking Institute} \\
	\textit{Carnegie Mellon University}\\
	Pittsburgh, PA 15213 \\
	xiaochux@alumni.cmu.edu}
}

\maketitle

\begin{abstract}
In this paper, we propose an optimized Transformer model that integrates Bayesian algorithms with a Bidirectional Gated Recurrent Unit (BiGRU), and apply it to fake news classification for the first time. First, we employ the TF-IDF method to extract features from news texts and transform them into numeric representations to facilitate subsequent machine learning tasks. Two sets of experiments are then conducted for fake news detection and classification: one using a Transformer model optimized only with BiGRU, and the other incorporating Bayesian algorithms into the BiGRU-based Transformer. Experimental results show that the BiGRU-optimized Transformer achieves 100\% accuracy on the training set and 99.67\% on the test set, while the addition of the Bayesian algorithm maintains 100\% accuracy on the training set and slightly improves test-set accuracy to 99.73\%. This indicates that the Bayesian algorithm boosts model accuracy by 0.06\%, further enhancing the detection capability for fake news. Moreover, the proposed algorithm converges rapidly at around the 10th training epoch with accuracy nearing 100\%, demonstrating both its effectiveness and its fast classification ability. Overall, the optimized Transformer model, enhanced by the Bayesian algorithm and BiGRU, exhibits excellent continuous learning and detection performance, offering a robust technical means to combat the spread of fake news in the current era of information overload.
\end{abstract}

\begin{IEEEkeywords}
Bayesian algorithm; fake news detection; transformer; BiGRU.
\end{IEEEkeywords}

\section{Introduction}

\IEEEPARstart{T}{he} rapid expansion of the Internet and social media has significantly accelerated the spread of fake news, posing serious challenges across social, political, and economic domains~\cite{1}. Defined as misleading or fabricated content designed to attract attention, manipulate opinions, or serve specific agendas, fake news propagates rapidly through digital communication networks, often leading to misinformation crises. For instance, during elections, fake news influences voter decision-making and undermines democratic integrity. Consequently, automated fake news detection has become a critical research focus in information science, data science, and computational social science~\cite{2}.

Among the various approaches explored, machine learning (ML) has emerged as a key solution, enabling automated analysis and classification of misinformation. ML models extract distinguishing features from historical and spatiotemporal data~\cite{3}, including linguistic patterns (e.g., sentiment analysis, word frequency), user interaction metrics (e.g., engagement levels, virality), and source credibility~\cite{4}. Common ML-based classifiers include support vector machines (SVM) , decision trees , random forests, and neural networks, while deep learning architectures such as convolutional neural networks (CNN), recurrent neural networks (RNN), RoBERTa, DeBERTa, and T5 have demonstrated superior performance in handling complex textual data~\cite{5}. Compared to traditional approaches, these models offer enhanced accuracy and robustness in identifying misinformation.

To address the data scarcity challenge in fake news detection, semi-supervised learning and transfer learning techniques have been employed to leverage both labeled and unlabeled data. Additionally, recent advancements in large language models (LLMs) have significantly improved detection capabilities by integrating multimodal learning, adversarial training, and chain of reasoning~\cite{6,7,18,20}. Retrieval-Augmented Generation (RAG) was proposed for further improve performance of LLMs.~\cite{8}. Pre-trained LLMs, for example, GreenPLM, not only showed great performance but also had low cost for training.~\cite{9} However, challenges remain in adapting to evolving misinformation trends and avoiding uncertainty of Large Language Models~\cite{10}.

In this paper, we propose an optimized Transformer-based model, incorporating Bayesian inference and bidirectional gated recurrent units (Bi-GRUs) to enhance fake news classification accuracy. To the best of our knowledge, this is the first application of this approach in misinformation detection~\cite{19}.

\section{Data Sources}

The data set selected in this paper comes from the Kaggle open source data set, which contains 5000 rows of data, including two categories of true news and fake news. The data set has been tested by numerous experimenters in kaggle, and can significantly distinguish and compare the advantages and disadvantages of the algorithm. Select some data sets for display, and the results are shown in Table~\ref{tab:1}.

\begin{table}[!ht]
	\centering
	\caption{Some of the data} \label{tab:1}
	\begin{tabular}{m{7.5cm}<{\centering}c}
		\hline
		Text & Type  \\
		\hline
		Trump says healthcare reform push may need additional money WASHINGTON (Reuters) - President Donald Trump on Tuesday said that the Republican push to repeal Obamacare may require additional money for healthcare, but he did not specify how much more funding would be needed or how it might be used. Trump told Republican Senators joining him for lunch at the White House that their planned healthcare reform bill would need to be ``generous'' and ``kind.'' ``That may be adding additional money into it,'' Trump said, without offering further details.~\cite{11}  &  Real \\
		\hline
		China's Xi, Trump discuss `global hot-spot issues': Xinhua BEIJING (Reuters) - Chinese President Xi Jinping and U.S. President Donald Trump on Saturday discussed ``global hot-spot issues'' on the sidelines of the G20 summit in the German city of Hamburg, state news agency Xinhua said. It did not immediately give any other details.  &  Real \\
		\hline
		Trump has talked to top lawmakers about immigration reform: White House WASHINGTON (Reuters) - U.S. President Donald Trump has spoken to congressional leaders about immigration reform and is confident that Congress will take action to deal with the status of illegal immigrants who have grown up in the United States, the White House said on Tuesday. ``We have confidence that Congress is going to step up and do their job,'' White House spokeswoman Sarah Sanders told a briefing shortly after the administration scrapped a program that protected from deportation some 800,000 young people who grew up in the United States. ``This is something that needs to be fixed legislatively and we have confidence that they're going to do that,'' Sanders said, adding that Trump was willing to work with lawmakers on immigration reform, which she said should include several ``big fixes,'' not just one tweak to the system.  &  Real \\
		\hline
	\end{tabular}
\end{table}

\section{Text Feature Extraction}

Term Frequency-inverse Document Frequency (TF-IDF) is probably the most common feature extraction method applied to text features in both Natural Language Processing and Information Retrieval. TF-IDF attempts to provide a measure of importance of a word in a particular document in context of its universality across the document set. More precisely, TF (word frequency) calculates the frequency of a word within a document, while IDF (inverse document frequency) calculates the rarity of a word across the total set of documents~\cite{12}. The score in TF-IDF comes to some value for every word just by multiplying both and may show how relevant each word is in that given document. This process includes data preprocessing (such as word segmentation, removal of stop words and desiccation), then calculating word frequency and inverse document frequency for each document, and finally obtaining a sparse matrix representing the TF-IDF value of each word in all documents.~\cite{13} These values are used as input features of subsequent machine learning models. The features converted to numerical types are used for subsequent machine learning classification.

\section{Method}

\subsection{Bayesian algorithm}

\begin{figure}[!h]
	\centering
	\includegraphics[width=0.6\linewidth]{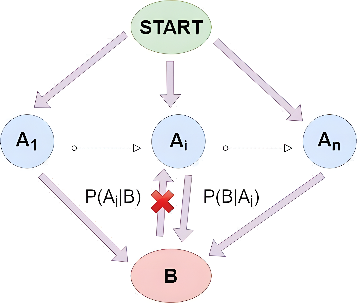}
	\caption{The principle diagram of Bayes algorithm.}
	\label{fig:1}
\end{figure}

Bayesian algorithm is a statistical inference method based on Bayes' theorem for classification and probabilistic model construction. The core idea is to evaluate the probability of an event by updating the prior information and combining the new observation data. The principle diagram of Bayes algorithm is shown in Fig.~\ref{fig:1}. The Bayesian approach, which emphasizes adjusting our beliefs by observing new evidence during reasoning, is flexible and efficient.

In a Bayesian framework, we usually start with prior knowledge, which may be derived from historical data or the experience of domain experts. A prior probability quantifies our initial belief that an event will occur in the absence of observational data. When new observational data appear, Bayesian algorithms are used to update this belief, producing a posterior probability. This updating process attaches importance to the information provided by the data, which means that even if the prior knowledge is poor, the prediction ability of the model will gradually improve with the addition of more data~\cite{14}.

The Bayesian methods of reasoning specifically have to deal with problems having uncertainty and complexities. Most practical problems involve usually incomplete knowledge of existing and observational data. The Bayesian algorithms can then become very strong, using prior distributions in a manner such that small data may provide a very accurate maintenance of performance. The flexibility makes Bayesian methods bound for application in domains of wide variance, including but not limited to medicine, finance, machine learning, and natural language processing.

\subsection{Bidirectional gated cycle unit}

\begin{figure}[!h]
	\centering
	\includegraphics[width=\linewidth]{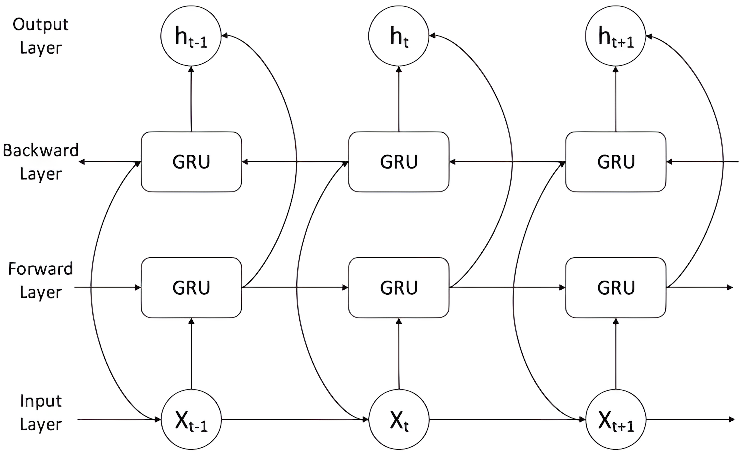}
	\caption{The schematic diagram of the bidirectional gated cycle unit.}
	\label{fig:2}
\end{figure}

Bi-gated loop Unit (Bi-GRU) is an improved recurrent neural network (RNN) structure for the processing of sequence data, such as natural language processing and time series prediction. Fig.~\ref{fig:2} shows the schematic diagram of bidirectional gated cycle unit. Unlike traditional one-way recurrent neural networks, the bidirectional gated recurrent structure improves the understanding of the model in context by considering both the forward and reverse information of the sequence~\cite{15}.

The working mechanism of bidirectional GRUs is divided into two major parts: forward and reverse. In the forward part, the model is designed to take input data in steps in a normal chronological manner, passing information from beginning to end. In the backward part, it models the information from the end to the beginning to capture the very ending information that influences the overall context. By this means, the bidirectional flow of information improves the ability of the model to capture deeper relationships of contexts and enriches the ability of extracting features in a sequence. 

GRU itself is a variant of the RNN, which was proposed to alleviate the problem of disappearing gradients while doing long-term dependency learning for traditional RNNS. Unlike traditional RNNS, GRUs guide the flow of information by introducing a gating mechanism that makes decisions on what it needs to retain and what should be forgotten; hence, this makes the GRU much more efficient and robust in learning the long sequences. Bidirectional GRU, at each step, goes both forward and backward to provide the last state of output; hence, bidirectional GRU has some advantages over a standard one during decision making within a larger context~\cite{16}.

Bi-directional GRUs do find an extremely important application in NLP domain tasks, such as Language Modeling, Sentiment Analysis, and Machine Translation, where it has become usual to capture the context with meanings on both sides around certain phrases to decide the meaning of a phrase as a sentence wholesaler. By modeling contextual relationships in both directions of context simultaneously, it allows for more accurate modeling of text, which is useful in the performance of such a task.

\section{Transformer}

\begin{figure}[!h]
	\centering
	\includegraphics[width=\linewidth]{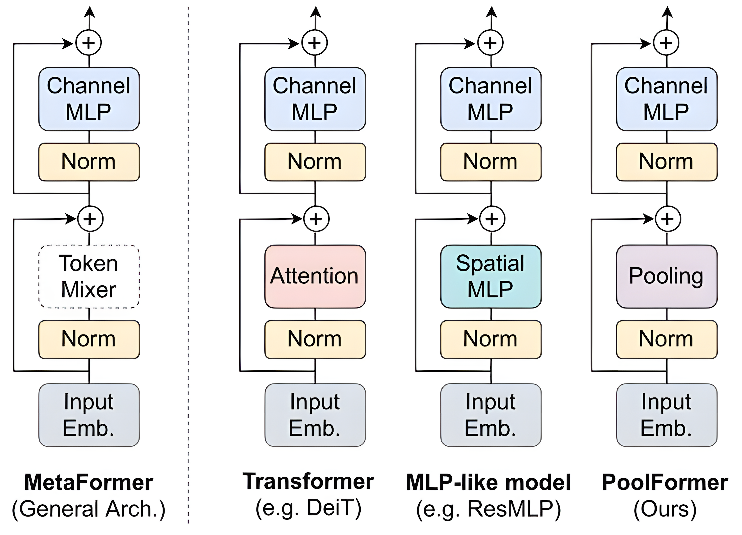}
	\caption{A schematic diagram of Transformer.}
	\label{fig:3}
\end{figure}

Transformer is a deep learning model for processing sequence data, first proposed in the year 2017 by Vaswani et al. Since its proposition, it has entirely changed model design in the field of natural language processing (NLP), especially on machine translation tasks. Unlike traditional recurrent neural networks (RNN), Transformer is completely based on mechanisms of self-attention and abandons series-dependent limitations~\cite{17,19}. This makes parallel processing possible and significantly improves training efficiency. A schematic diagram of Transformer is presented in Fig.~\ref{fig:3}.

The self-attention mechanism forms the core of the Transformer model for its ability to capture correlations between positions in an input sequence. Computing similarities for a word against all other words in the sequence, it assigns attention weights variably, helping it catch contextual information effectively. This approach effectively helps in capturing long-range dependencies because every word can directly interact with every other word in the sequence, without necessarily relying on sequential layers to pass information.

The two major components of the Transformer model are the encoder and decoder. The encoder mainly translates the input sequence into higher levels of context-based representations, which are then used by the decoder for generating the target output sequence. The encoder consists of a stack of identical layers, each having a multi-head self-attention mechanism and a feed-forward neural network as two major components. This is the architecture that results in parallelization along the processing of multiple representation subspaces, which intrinsically improves generalization capability and expressiveness.

In some ways, the architecture of decoder is related to that of the encoder with an added manner of autoregressive generation, which means for every step it generates, there is dependence from previously generated ones. Multi-headed self-attention plays an essential role in achieving this by considering all previously created words and encompassing information emanating from all the encoder outputs at each individual generation step.

Another important feature in the Transformer model is positional encoding. Since the self-attention mechanism itself doesn't capture  position information of words in a sequence, positional encoding introduces either relative or absolute position information. This would allow the model to know that the input comes in some sort of sequence, which it could keep in word order in its representations.

\subsection{Transformer algorithm based on Bayesian algorithm and bidirectional gated cycle unit optimization}

\begin{figure}[!h]
	\centering
	\includegraphics[width=0.7\linewidth]{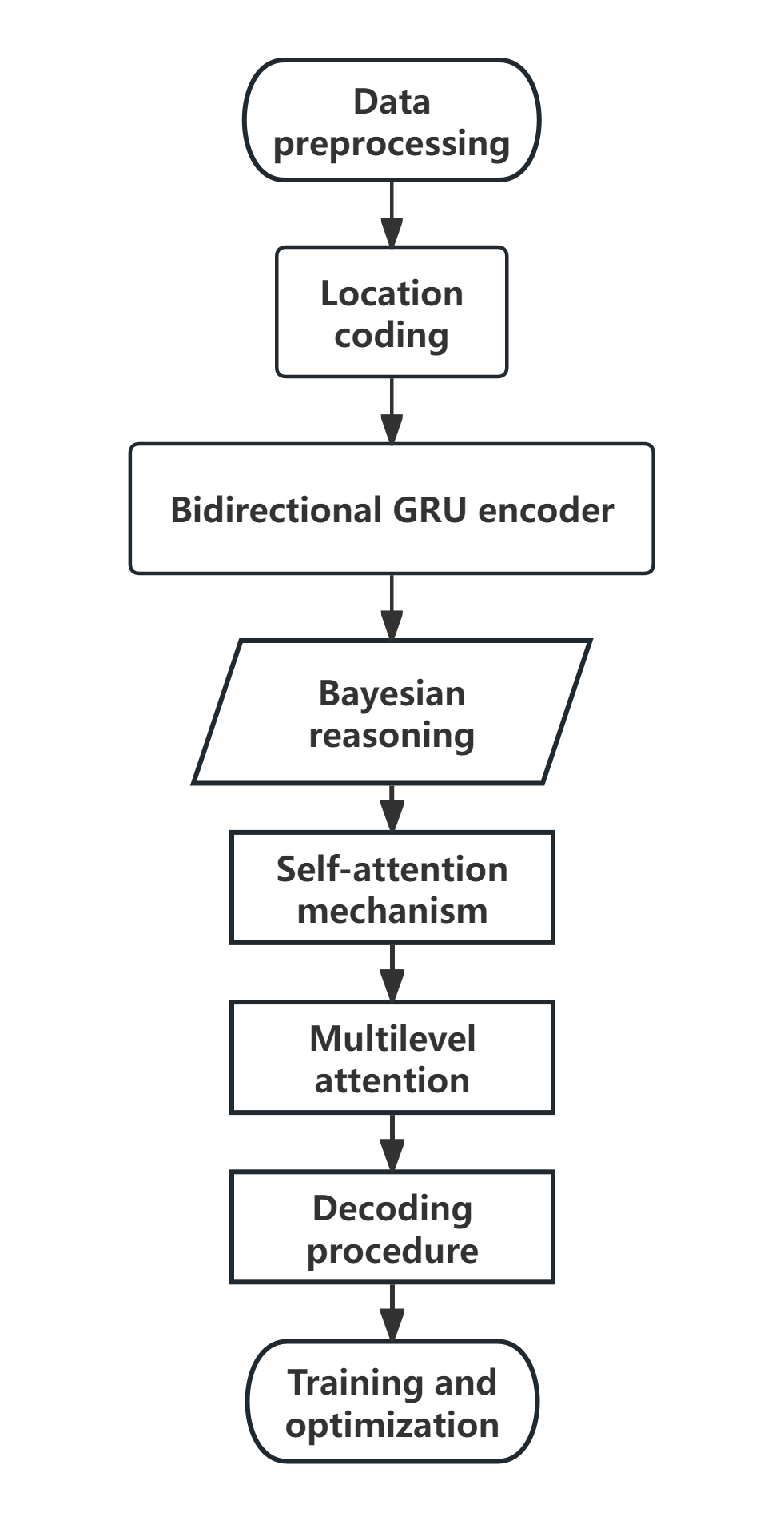}
	\caption{The working flow chart of the algorithm.}
	\label{fig:4}
\end{figure}

The Transformer algorithm, which integrates Bayesian inference and Bidirectional Gated Recurrent Unit (Bi-GRU) optimization, enhances the performance and uncertainty quantification of models processing sequential data by combining self-attention mechanisms in deep learning with Bayesian inference. By incorporating Bi-GRU into the Transformer's encoding and decoding modules, the model can capture contextual information from both preceding and succeeding elements. Concurrently, the application of Bayesian algorithms for parameter inference allows the model to effectively manage uncertainties and improve generalization capabilities. This results in higher accuracy and robustness in tasks such as natural language processing and sequence prediction. The workflow of the algorithm is illustrated in Fig.~\ref{fig:4}.

Bayesian algorithm is mainly used to optimize the uncertainty estimation of model parameters. By introducing Bayesian inference, the model is able to model the probability distribution of the parameters rather than a single point estimate. Specifically, the Bayesian algorithm computes the posterior distribution of the parameters through the prior distribution and the likelihood function of the observed data, thereby dynamically adjusting the uncertainty of the parameters during training. This uncertainty estimation helps to improve the robustness of the model, especially when the data is sparse or noisy. Combined with Bayesian optimization, Transformer can better capture potential patterns in input sequences while reducing the risk of overfitting and improving generalization.

\subsubsection{Data preprocessing} 

The input sequence is preprocessed as necessary, including word segmentation, conversion to vector representation, and padding and truncation to fixed length.

\subsubsection{Location coding} Since Transformer cannot handle sequence locations directly, add location encoding to retain sequence information.

\subsubsection{Bidirectional GRU encoder} 

The input sequence is encoded using a bidirectional GRU. The bidirectional GRU reads the entire sequence, taking the context representation in both the forward and backward directions, and generating hidden states.

\subsubsection{Bayesian reasoning}

Based on the generated hidden state, Bayesian inference is used to update the model parameters or generate a probability distribution for the sequence. With a Bayesian approach, the prior distribution can be integrated and the posterior distribution updated based on the input data, which gives the model greater reasoning power.

\subsubsection{Self-attention mechanism}

In Transformer, the self-attention mechanism is used to calculate the attention weight of the input sequence. This step combines the contextual information obtained by BiGRU with the representation of each word to get richer information.

\subsubsection{Multiple layers of attention}
 
On the basis of self-attention, multiple attention layers are constructed to focus on different parts of the sequence at the same time through multiple independent attention mechanisms. Each head learns a different representation, enhancing the power of the model.

\subsubsection{Feedforward neural network}

The output of multi-head attention is passed into the feedforward neural network to further process the information and introduce the nonlinearity through the activation function.

\subsubsection{Decode}

In the decoding phase, the output is generated sequentially, relying on the autoregressive model structure, while taking into account the contextual information provided by the encoder and the previously generated output.

\subsubsection{Training and optimization}

The model is trained by gradient descent optimization algorithm to minimize the loss function. At the same time, the parameters are updated by Bayesian inference to ensure that the model can still make correct predictions in the case of high uncertainty.

\section{Result}

In terms of parameter Settings, Adam Optimizer was used in the experiment, the maximum training rounds is set to 200, the number of batches is set to 256, the initial learning rate is 0.001, the learning rate decline factor is set to 0.1, and the gradient clipping threshold is set to 10. The Nvidia 4090 GPU is used for running experiments with Matlab R2024a.

In the division of data sets, this experiment divided the training set and the testing set according to the ratio of 7:3. In the category of binary classification, the proportion of data sets is balanced.

\begin{figure}[!h]
	\centering
	\includegraphics[width=0.6\linewidth]{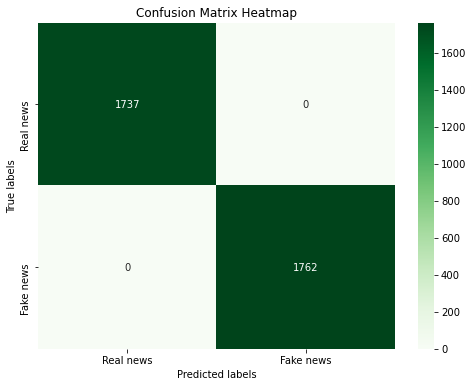}
	\includegraphics[width=0.6\linewidth]{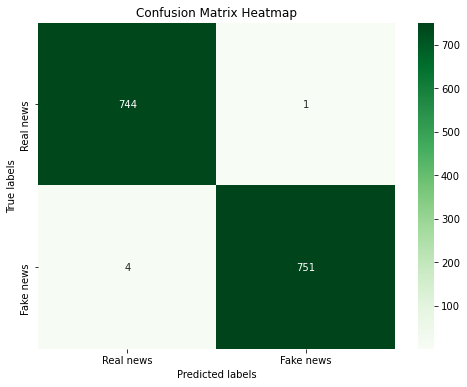}
	\caption{The experiment of fake news detection and classification with the Transformer algorithm based on bidirectional gated cycle unit optimization.}
	\label{fig:5}
\end{figure}

\begin{figure}[!h]
	\centering
	\includegraphics[width=0.6\linewidth]{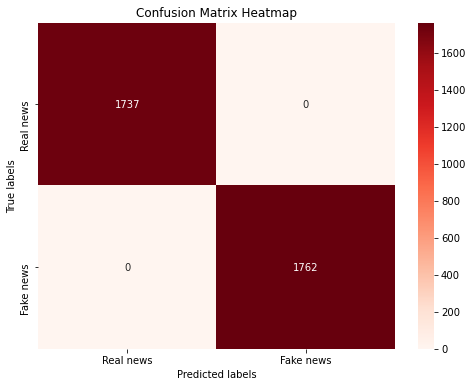}
	\includegraphics[width=0.6\linewidth]{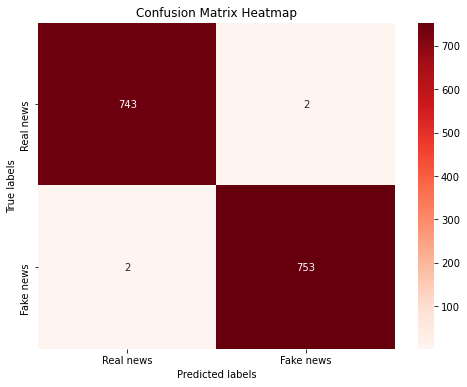}
	\caption{The experiment of fake news detection and classification based on Bayesian algorithm and bidirectional gated cycle unit optimization Transformer algorithm.}
	\label{fig:6}
\end{figure}

This paper presents experiments using two variations of the Transformer algorithm: one optimized with a bidirectional gated cycle unit and another optimized with a bidirectional gated cycle unit based on Bayesian optimization. To compare their performance, we analyze the confusion matrices generated for both the training and testing datasets. Fig.~\ref{fig:5} shows the confusion matrix of the Transformer model optimized with a bidirectional gated cycle unit, while Fig.~\ref{fig:6} shows the confusion matrix of the model with Bayesian optimization and the bidirectional gated cycle unit.

In the experiment of fake news detection with the Transformer algorithm based on bidirectional gated cycle unit optimization, the accuracy of the training set is 100\%, and the accuracy of the testing set is 99.67\%.

In the experiment of fake news detection based on Bayesian algorithm and bidirectional gated cycle unit optimization Transformer algorithm, the accuracy of the training set is 100\%, and the accuracy of the test set is 99.73\%.

The accuracy of both algorithms on the training and testing sets is compared and presented in Table~\ref{tab:2}.

\begin{table}[!ht]
	\centering
	\caption{The accuracy of two algorithms on training set and testing set} \label{tab:2}
	\begin{tabular}{m{5.5cm}<{\centering}cc}
		\hline
		Method & \makecell{Training\\ accuracy\\ (\%)}  & \makecell{Training\\ accuracy\\ (\%)} \\
		\hline
		Bidirectional Gated Cycle Unit Optimization Transformer & 100 & 99.67 \\
		Bidirectional gated Cycle unit with Bayesian Algorithm Optimization Transformer & 100 & 99.73 \\
		\hline
	\end{tabular}
\end{table}

According to the experimental results, the accuracy rate of the two algorithms on the training set is 100\%, and the classification performance of two algorithms on the testing set is more than 99\%, showing great fake news detection and classification accuracy. In addition, after adding Bayesian algorithm optimization, the accuracy of the Transformer algorithm based on bidirectional gated cycle unit optimization increases by 0.06\%, which further improves the prediction accuracy.

\begin{figure}[!h]
	\centering
	\includegraphics[width=\linewidth]{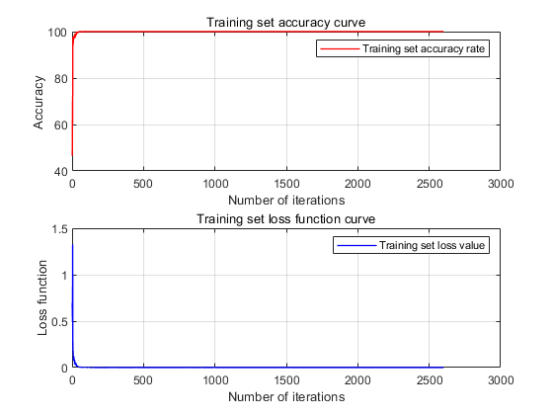}
	\caption{The change curve of loss and accuracy of the algorithm.}
	\label{fig:7}
\end{figure}

Fig.~\ref{fig:7} shows the loss and accuracy curves of the algorithm. From the curves, we found that the algorithm used in this experiment converges around the 10th epoch, achieving nearly 100\% accuracy. This indicates that the algorithm have excellent classification speed, allowing it to quickly and accurately detect and classify fake news.

\section{Conclusion}

This study is the first to apply Bayesian algorithm and Bidirectional gated loop unit (Bi-GRU) optimization Settings to the classification and prediction task of fake news. We utilized TF-IDF method to extract features from news texts and convert them into numerical features used by subsequent machine learning models. In the experiment, we designed two different models: Transformer algorithm based on Bi-GRU optimization, and Transformer algorithm combining Bayes algorithm and Bi-GRU optimization. The results showed that both algorithms achieve great performance in detecting fake news. When using a single Bi-GRU optimized Transformer algorithm, the accuracy of the training set reached 100\%, while the accuracy of the testing set was 99.67\%. The optimized version with the addition of the Bayesian algorithm also achieved 100\% accuracy on the training set and increased to 99.73\% on the testing set. This increased accuracy of 0.06\% demonstrated the Bayesian algorithm as an important component for optimizing the Transformer model. 

Analyzing the training accuracy and loss function curves shows that the model is close to convergence around the 10th epoch, and the accuracy is close to 100\%. This indicates that the proposed optimization algorithm not only has fast classification speed, but also can achieve high-precision for fake news detection in a short time. This point has important practical applications for the early warning and accurate identification of fake news.

In conclusion, this study proves the effectiveness and performance of Transformer algorithm based on the Bayesian algorithm and Bi-GRU optimization in fake news classification problem. Such combinations not only improve the classification accuracy, but also provide a new solution to the growing challenge of false information. This achievement provides important reference value and inspiration for future related research.

\bibliographystyle{IEEEtran}
% \bibliography{paper}

\begin{thebibliography}{10}
\providecommand{\url}[1]{#1}
\csname url@samestyle\endcsname
\providecommand{\newblock}{\relax}
\providecommand{\bibinfo}[2]{#2}
\providecommand{\BIBentrySTDinterwordspacing}{\spaceskip=0pt\relax}
\providecommand{\BIBentryALTinterwordstretchfactor}{4}
\providecommand{\BIBentryALTinterwordspacing}{\spaceskip=\fontdimen2\font plus
\BIBentryALTinterwordstretchfactor\fontdimen3\font minus \fontdimen4\font\relax}
\providecommand{\BIBforeignlanguage}[2]{{%
\expandafter\ifx\csname l@#1\endcsname\relax
\typeout{** WARNING: IEEEtran.bst: No hyphenation pattern has been}%
\typeout{** loaded for the language `#1'. Using the pattern for}%
\typeout{** the default language instead.}%
\else
\language=\csname l@#1\endcsname
\fi
#2}}
\providecommand{\BIBdecl}{\relax}
\BIBdecl

\bibitem{1}
C.-M. Lai, M.-H. Chen, E.~Kristiani, V.~K. Verma, and C.-T. Yang, ``Fake news classification based on content level features,'' \emph{Applied Sciences}, vol.~12, no.~3, p. 1116, 2022.

\bibitem{2}
D.~Rohera, H.~Shethna, K.~Patel, U.~Thakker, S.~Tanwar, R.~Gupta, W.-C. Hong, and R.~Sharma, ``A taxonomy of fake news classification techniques: Survey and implementation aspects,'' \emph{IEEE Access}, vol.~10, pp. 30\,367--30\,394, 2022.

\bibitem{3}
Z.~Zhang, X.~Wang, X.~Zhang, and J.~Zhang, ``Simultaneously detecting spatiotemporal changes with penalized poisson regression models,'' \emph{arXiv preprint arXiv:2405.06613}, 2024.

\bibitem{4}
X.~Xu, P.~Yu, Z.~Xu, and J.~Wang, ``A hybrid attention framework for fake news detection with large language models,'' \emph{arXiv preprint arXiv:2501.11967}, 2025.

\bibitem{5}
X.~Huang, Y.~Wu, D.~Zhang, J.~Hu, and Y.~Long, ``Improving academic skills assessment with nlp and ensemble learning,'' in \emph{2024 IEEE 7th International Conference on Information Systems and Computer Aided Education (ICISCAE)}.\hskip 1em plus 0.5em minus 0.4em\relax IEEE, 2024, pp. 37--41.

\bibitem{6}
J.~Yi, Z.~Xu, T.~Huang, and P.~Yu, ``Challenges and innovations in llm-powered fake news detection: A synthesis of approaches and future directions,'' \emph{arXiv preprint arXiv:2502.00339}, 2025.

\bibitem{7}
J.~He, M.~D. Ma, J.~Fan, D.~Roth, W.~Wang, and A.~Ribeiro, ``Give: Structured reasoning with knowledge graph inspired veracity extrapolation,'' \emph{arXiv preprint arXiv:2410.08475}, 2024.

\bibitem{18}
H.~Guo, T.~Huang, H.~Huang, M.~Fan, and G.~Friedland, ``A systematic review of multimodal approaches to online misinformation detection,'' in \emph{2022 IEEE 5th International Conference on Multimedia Information Processing and Retrieval (MIPR)}.\hskip 1em plus 0.5em minus 0.4em\relax IEEE, 2022, pp. 312--317.

\bibitem{20}
P.~Yu, X.~Xu, and J.~Wang, ``Applications of large language models in multimodal learning,'' \emph{Journal of Computer Technology and Applied Mathematics}, vol.~1, no.~4, pp. 108--116, 2024.

\bibitem{8}
W.~Liu, J.~Chen, K.~Ji, L.~Zhou, W.~Chen, and B.~Wang, ``Rag-instruct: Boosting llms with diverse retrieval-augmented instructions,'' \emph{arXiv preprint arXiv:2501.00353}, 2024.

\bibitem{9}
Q.~Zeng, L.~Garay, P.~Zhou, D.~Chong, Y.~Hua, J.~Wu, Y.~Pan, H.~Zhou, R.~Voigt, and J.~Yang, ``Greenplm: cross-lingual transfer of monolingual pre-trained language models at almost no cost,'' \emph{arXiv preprint arXiv:2211.06993}, 2022.

\bibitem{10}
Q.~Zeng, M.~Jin, Q.~Yu, Z.~Wang, W.~Hua, Z.~Zhou, G.~Sun, Y.~Meng, S.~Ma, Q.~Wang \emph{et~al.}, ``Uncertainty is fragile: Manipulating uncertainty in large language models,'' \emph{arXiv preprint arXiv:2407.11282}, 2024.

\bibitem{19}
H.~Guo, T.~Huang, H.~Huang, M.~Fan, and G.~Friedland, ``Detecting covid-19 conspiracy theories with transformers and tf-idf,'' \emph{arXiv preprint arXiv:2205.00377}, 2022.

\bibitem{11}
W.~H. Bangyal, R.~Qasim, N.~U. Rehman, Z.~Ahmad, H.~Dar, L.~Rukhsar, Z.~Aman, and J.~Ahmad, ``Detection of fake news text classification on covid-19 using deep learning approaches,'' \emph{Computational and mathematical methods in medicine}, vol. 2021, no.~1, p. 5514220, 2021.

\bibitem{12}
L.~Bozarth and C.~Budak, ``Toward a better performance evaluation framework for fake news classification,'' in \emph{Proceedings of the international AAAI conference on web and social media}, vol.~14, 2020, pp. 60--71.

\bibitem{13}
W.~Liu, S.~Cheng, D.~Zeng, and H.~Qu, ``Enhancing document-level event argument extraction with contextual clues and role relevance,'' \emph{arXiv preprint arXiv:2310.05991}, 2023.

\bibitem{14}
M.~Z. Nawaz, M.~S. Nawaz, P.~Fournier-Viger, and Y.~He, ``Analysis and classification of fake news using sequential pattern mining,'' \emph{Big Data Mining and Analytics}, vol.~7, no.~3, pp. 942--963, 2024.

\bibitem{15}
N.~Rai, D.~Kumar, N.~Kaushik, C.~Raj, and A.~Ali, ``Fake news classification using transformer based enhanced lstm and bert,'' \emph{International Journal of Cognitive Computing in Engineering}, vol.~3, pp. 98--105, 2022.

\bibitem{16}
M.~Fayaz, A.~Khan, M.~Bilal, and S.~U. Khan, ``Machine learning for fake news classification with optimal feature selection,'' \emph{Soft Computing}, vol.~26, no.~16, pp. 7763--7771, 2022.

\bibitem{17}
P.~Yu, J.~Yi, T.~Huang, Z.~Xu, and X.~Xu, ``Optimization of transformer heart disease prediction model based on particle swarm optimization algorithm,'' \emph{arXiv preprint arXiv:2412.02801}, 2024.

\end{thebibliography}

\end{document}